# Decoding excellence: Mapping the demand for psychological traits of operations and supply chain professionals through text mining

Sebastiano Di Luozzo [1], Andrea Fronzetti Colladon [2], Massimiliano M. Schiraldi [1]

[1] *University of Rome Tor Vergata*

[2] *University of Perugia*

***Abstract***

The current study proposes an innovative methodology for the profiling of psychological traits of Operations Management (OM) and Supply Chain Management (SCM) professionals. We use innovative methods and tools of text mining and social network analysis to map the demand for relevant skills from a set of job descriptions, with a focus on psychological characteristics. The proposed approach aims to evaluate the market demand for specific traits by combining relevant psychological constructs, text mining techniques, and an innovative measure, namely, the Semantic Brand Score. We apply the proposed methodology to a dataset of job descriptions for OM and SCM professionals, with the objective of providing a mapping of their relevant required skills, including psychological characteristics. In addition, the analysis is then detailed by considering the region of the organization that issues the job description, its organizational size, and the seniority level of the open position in order to understand their nuances. Finally, topic modeling is used to examine key components and their relative significance in job descriptions. By employing a novel methodology and considering contextual factors, we provide an innovative understanding of the attitudinal traits that differentiate professionals. This research contributes to talent management, recruitment practices, and professional development initiatives, since it provides new figures and perspectives to improve the effectiveness and success of Operations Management and Supply Chain Management professionals.

***Keywords***

Human Factor, Text Mining, Semantic Brand Score, Semantic Network Analysis, IPIP, Operations and Supply Chain Management

# 1 Introduction

Analytical modeling is not sufficient to predict operations and supply chain outcomes. As a result, research in operations management (OM) and supply chain management (SCM) is changing due to the integration of behavioral and cognitive aspects into classical models (Bendoly et al. 2006). Behavioral operations theory emerged from these considerations (Gino and Pisano 2008a) and has rapidly gained traction in recent years (Donohue et al. 2019)[1]. The recognition of psychology's crucial role in building effective OM and SCM frameworks has been the first step toward the field's remarkable growth (Boudreau et al. 2003, McClain et al. 1992, Ostolaza and McClain 1990, Van Oyen et al. 2001), which is now consistently reinforced by the many scientific findings obtained within the research area. For example, one of the most relevant findings is given by evaluating the well-known "Newsvendor problem" from a behavioral point of view (Long and Nasiry 2014, Ren and Croson 2013). When incorporating behavioral considerations into the traditional model, it appears evident that the solutions typically differ from the optimal (Ho et al. 2010, Schweitzer and Cachon 2000) with the assumption of human rationality. As a result of many years of research, it is now clear that classical assumptions (Bendoly et al. 2006) – such as people being deterministic and predictable in their actions, emotionless, and observable – lead to flawed findings and ineffective OM and SCM frameworks (Tokar 2010), hence requiring the integration of human behavior within the operations and supply chain theory.

Given this landscape, the psychological mapping and evaluation of individuals' attitudes – and, more generally, the human factor – for recruiting and organizational development purposes appears to be one of the most promising areas to reap the benefits of behavioral operations research (Aloini et al. 2021, Gino and Pisano 2008). By better understanding the psychological and social dynamics underpinning OM and SCM actions, one can anticipate specific operational situations and their potential consequences on team and company performance (Bendoly et al. 2010), enabling individuals to receive individualized Human Resource Management (HRM) based on their characteristics (Aloini et al. 2021). Croson et al. (2013) also maintain this view, calling personnel assessment 'vital' for the behavioral operations theory and considering that non-standard measurement practices for the field of OM field (e.g., psychometric measures) may foster its development.

[1] According to the analysis of 58 published contributions in this field from 2000 to 2021 (see *Appendix A* for details on the research works). These references have been selected through the Knowledge Development Process-Constructivist methodology (Dutra et al., 2015), from Web of Science and Scopus databases through the following search query: [behavior* OR behaviour*] AND [operations OR supply chain].

In this scenario, the analysis of job descriptions provides practical insights into prominent individual characteristics required by organizations and allows for drawing up empirical findings from the current recruiting practices (Boudreau et al. 2003, Cohen et al. 2020, Dwivedi et al. 2020). As an indispensable tool in modern data analysis, text mining can contribute significantly to this endeavor by enabling the efficient processing of large volumes of textual data and the extraction of patterns. Following this consideration, some authors have started to adopt text mining techniques for the evaluation of relevant skills from data analysis (Abdelall et al. 2012a, Wagner et al. 2019a). An example is given by the manuscript of Rossetti and Dooley (2010), which performs an analysis of SCM job descriptions by implementing text mining algorithms to identify relevant job types and market requirements. However, only a few scientific contributions have attempted to provide these measurements, especially through implementing Natural Language Processing (NLP) techniques (Donohue et al. 2020, Fareri et al. 2021). Authors have often focused on competencies and hard skills rather than specifically on attitudes and personality traits, thus creating a gap in the behavioral operations research area. The identified research gap is further detailed as follows, along with the related research questions. Accordingly, this paper tries to answer the following research questions: *How can the demand for Operations and Supply Chain professionals be described from a psychological perspective, considering current recruitment practices and the scientific literature? Are there differences in relation to specific job or company characteristics (e.g., size of organization, etc.)?*

This work contributes to the field of behavioral operations research in two ways. First, we propose a methodology for the psychological mapping of professionals from job descriptions. The proposed approach combines relevant psychological constructs, NLP techniques, and innovative methods of semantic network analysis. Second, we apply the proposed methodology to a set of job descriptions for Operations and Supply Chain professionals to map the required skills and psychological characteristics.

The remainder of this paper is organized as follows. Section 2 introduces the conceptual framework for assessing the demand for psychological traits of professionals using NLP techniques. We contextualize this specific research area within the field of behavioral operations. Section 3 describes the proposed methodology in detail, and Section 4 applies it to a set of job descriptions. Section 5 discusses the most relevant results and compares them with insights from the scientific literature. We then offer some concluding remarks and describe our work's limitations and further development in Section 6.

## 2 Conceptual Framework

People represent the central element underlying operations and supply chain management contexts and activities (Croson et al. 2013), and it is no surprise that human capital is often referred to as the essential asset for an organization (Fareri et al. 2020). From the latter assumption, behavioral operations theory has originated and spread rapidly among researchers and practitioners. This traction has given birth to different research streams focused on the evaluation of operational performances based on individual aspects, such as reputation (Katok and Siemsen 2011), non-conventional behaviors (Bellezza et al. 2014), social preferences (Loch and Wu 2008) – or overall team attributes and organizational factors, such as level of coordination (Katok and Villa 2021), teamwork and cooperation (Lee et al. 2013), and interpersonal relations (Kashdan and Roberts 2004). Among those investigation areas, psychometric research is prominent but scarcely explored in the behavioral operations environment (Erjavec and Trkman 2020). Our manuscript addresses this gap by providing a conceptual framework for the development of the research field (see **Error! Reference source not found.**), showing the main elements for the potential implementation of a behavioral assessment of OM and SCM professionals.

**Figure 1:** A conceptual framework for psychometric research in operations and supply chain management

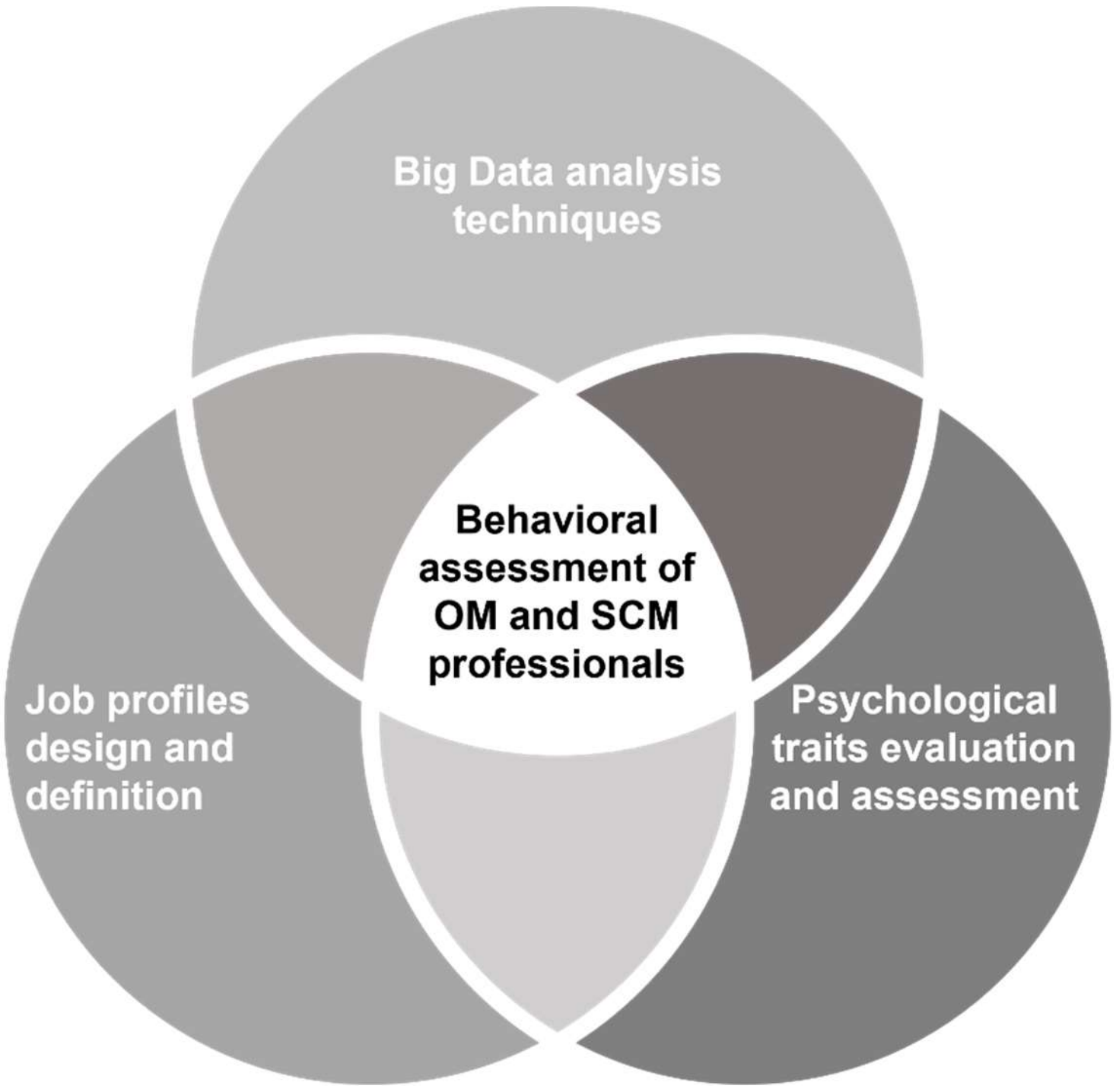

The theoretical model leans on three main pillars and is built on the use of big data techniques. According to Donohue et al. (2020), big data analysis techniques can play a crucial role in helping organizations valorize the individuals' and teams' aspects impacting operations and supply chain management outcomes (Aloini et al. 2021). To this end, given the advancements in technology and data management, adopting such capabilities allows direct observation of people's behaviors, drawing conclusions about factors influencing performance (Hazen et al. 2018). However, choosing a specific big data methodology depends on the required outcome and the research objective. Because our purpose is to determine the market demand for the main psychological traits characterizing professional roles in operations and supply chain management, job profiles represent a valuable source of data (Cohen et al. 2020, Rossetti and Dooley 2010) due to their availability, ease-of-use, and density of information to be collected. This opportunity is also allowed by the trend in recent years toward including soft skills, attitudinal qualifications, and noncognitive aspects in job descriptions rather than just listing technical or hard skills (Menon 2012). With this in mind, text mining exploiting Natural Language Processing (NLP) methodologies appears to be a suitable choice for the job description analysis due to their ability to extract relevant information from unstructured data sources, as shown by Fareri et al. (2021) and Almgerbi et al. (2021).

By employing this multifaceted conceptual framework that integrates a behavioral perspective with text mining and NLP techniques, we first contribute to advancing psychometric research in operations and supply chain management. Second, we provide a psychological mapping of the demand for leading professional figures in the field of research.

## 2.1 Psychological traits evaluation and assessment in Operations and Supply Chain Management

As personnel assessment and psychological traits evaluation have started becoming prominent research areas for the OM and SCM environment (Franco and Hämäläinen 2016, Zhao et al. 2013), the study of behavioral factors has allowed for a deeper analysis of classical models. Extending the OM and SCM frameworks into the behavioral domain is crucial for obtaining a comprehensive perspective on operational outcomes. However, it is not enough since it does not provide a holistic view of the factors underlying human choices. Instead, properly applying behavioral considerations to analytical models (Loch and Wu 2005, Tong and Feiler 2016) requires a far-reaching *mapping* of professionals' psychological traits.

The literature on psychological mapping of OM and SCM job figures is still in its early stages; however, notable contributions have paved the way for further studies with remarkable findings. An example is given by the case of buyer-supplier relationships, where research demonstrated that choices and decision-making processes are often influenced by personality disorders and psychopathy traits that executives and senior managers sometimes manifest (Croom et al. 2021). From a different perspective, Kalkanci et al. (2011) showed the operational effects on negotiations and contracts – e.g., longer negotiation process, less efficient contract, etc. – originating from a buyer-supplier relationship, which the personality disorders of managers may impact.

Inventory management is another well-investigated research stream from a behavioral point of view. Strohhecker and Größler (2013) noted that the traits associated with intelligence and knowledge are directly and positively linked with lower total inventory costs. On the contrary, openness to experience and extraversion seem to be negatively correlated with performance, leading to higher inventory costs. While some of these results may appear straightforward, they are not. Such outcomes describe in a simple fashion the psychological and scientific foundation underlying organizational performance and specifically one of the most well-known and relevant tasks for supply chain activities. This stream was analyzed in several other scientific contributions, investigating the influence of risk-averse behaviors (Zhang et al. 2018) and demand uncertainty (Ancarani et al. 2013) on inventory levels.

Following the previous considerations, as much work on personality traits in Operations and Supply Chain Management is focused on a limited number of profiles (Cohen et al. 2020, Dwivedi et al. 2020) or psychological factors (Pisano et al. 2016, Witt et al. 2017) or organizational outcomes (Liboni et al. 2019), further explorations are needed. Our research is not limited to the evaluation of one or more psychological traits in relation to a specific and known model or framework; it attempts to provide a general mapping of OM and SCM job profiles to establish a common foundation (Liu et al. 2019) for researchers and practitioners.

## 2.2 Natural Language Processing (NLP) techniques for skills and psychological traits assessment

The adoption of Natural Language Processing (NLP) techniques to assess individuals' skills and psychological traits has seen a significant increase in adoption in recent years (Donohue et al. 2020). Furthermore, the study of recent contributions in the fields of OM and SCM reveals a growing trend toward using NLP-based methods to extract information from unstructured data sources, such as text resumes, job descriptions, and social media profiles (Abdelall et al. 2012, Piselli et al. 2022, Wagner et al. 2019). This is confirmed by the work of Chiarello et al. (2021), which states the effectiveness of NLP-based methods in defining skills and psychological trait requirements from texts.

In this context, Fareri et al. (2020) introduced a Natural Language Processing methodology to assess specific skills within job profiles. The authors adopted an NLP-based technique to evaluate Industry 4.0's impact on job descriptions and validated their methodology with a real-world case study. Following a similar approach, the contribution of Almgerbi et al. (2021) used topic modeling to compare job descriptions and Massive Online Open Course (MOOC) programs, matching learning outcomes and job requirements for Information Technology (IT) professionals. These works provide relevant examples of the implementation of NLP techniques for skills mapping from text data sources that, however, do not perform any evaluation regarding the personality traits domain. Therefore, although NLP-based methods are gaining popularity and have proven their effectiveness in providing valuable insights into the qualifications and attributes of job applicants, there is no valid example of using such methods to assess individual personality traits. Taking this perspective into account and acknowledging the current scope of the literature, this manuscript aims to offer a contribution by proposing an NLP methodology that could be used in mapping individuals' personality traits. To the best of our knowledge, this approach is an initial step in exploring such applications within this domain.

## 2.3 Skills and psychological traits assessment frameworks

A categorization of soft skills and psychological traits is especially relevant when evaluating and assessing professional role profiles. The findings of numerous scientific investigations confirm this suggestion by indicating a positive correlation between organizational outcomes and soft skills, as well as specific personality traits that are sought after by managers and recruiters when making hiring decisions for job positions. Choi et al. (2020) highlighted the role of trust in supply chain decisions and its potential advantages in terms of business results. A similar result, yet from a different perspective, is obtained by Bellezza et al. (2014), who found that people who adopt nonconventional behaviors and can think out-of-the-box maximize the probability of being noticed, and this is positively related to superior organizational performances (Staats and Gino 2012). Although these contributions represent only a few examples of research demonstrating the importance of individuals' traits in operations, they reinforce the compelling need to consider aspects related to the human sphere in organizational decision-making processes. For this reason, taxonomies for skills and psychological traits assessment emerged in recent years.

In this context, the work of Stek and Schiele (2021) introduced a classification of the skills of Purchasing and Supply Management (PSM) professionals through a large survey conducted at the European level. The authors identified fifteen skill factors (e.g., orientation to results, imagination, etc.). They analyzed them according to specific organizational, purchasing, and supply objectives to determine the most relevant ones for operational results. Some of these factors are shared with the framework for attaining workforce agility proposed by Qin et al. (2015), such as the ability to adapt to changes. A different approach – even if with an analogous objective – is adopted by Bohlouli et al. (2017), who proposed a mathematical methodology to assess competencies by implementing a competence model measuring professional, innovation, and social characteristics.

Furthermore, since the classification of skills has been a critical component of the standardization of job profiles, efforts have been put toward their large-scale categorization and analysis by government organizations. The United States Department of Labor Dictionary of Occupational Titles (DOT)[2] – later replaced by the O*NET[3] framework – and the International Labour Organization's International Standard Classification of Occupations (ISCO)[4] are two of the earliest frameworks developed to facilitate this process. On a European level, the European Skills, Competences, Qualifications and Occupation (ESCO) project is one of the most comprehensive and effective frameworks for classifying skills and competencies. Like O*NET and ISCO, the ESCO database is a system for identifying and classifying relevant skills, competences, qualifications, and occupations in the European labor market. However, although the frameworks have been instrumental in providing a comprehensive understanding of the various aspects required for different job occupations, they do not have a specific focus on soft aspects and often neglect models concerning individuals' attitudes and psychological traits, such as the PPIK model theory (Ackerman 1996), the abridged Big Five-Dimensional Circumplex Mode (Bucher and Samuel 2019) or the NEO Personality Inventory (Costa and Mccrae 1992).

A major breakthrough in the context of personality assessments was obtained at the end of the '90s, when Goldberg introduced the International Personality Item Pool (IPIP) to enhance and encourage the diffusion of personality assessments (Johnson 2014). IPIP represents a freely accessible resource that collects over 4,000 items and 250 psychological constructs (http://ipip.ori.org/) that have been carefully selected based on a thorough review of the personality psychology literature and have been used in numerous empirical studies and large-scale assessments (Goldberg et al. 2006). The IPIP items are designed to be used with self-report questionnaires and are available for use in research studies and industrial applications, both within and outside academic settings (Donnellan et al. 2006). The key strength of the IPIP is its rigorous development process, which ensures the validity and reliability of its items (Goldberg 1999) by involving a consensus of experts in the field of personality psychology (Goldberg et al. 2006), helping to ensure that the IPIP items accurately capture individual differences in personality (Van Iddekinge et al. 2011, Ringwald et al. 2023).

Due to its characteristics, IPIP has been used to assess a wide range of personality dimensions relevant to job performance and success. For example, research has demonstrated that personality traits such as agreeableness, conscientiousness, emotional stability, and openness to experience positively relate to job satisfaction and performance

[2] https://occupationalinfo.org/
[3] https://www.onetonline.org/
[4] https://www.ilo.org/public/english/bureau/stat/isco/

(Barrick and Mount 1991, Judge and Bono 2001). IPIP provides measures of these and other important dimensions of personality, making it an effective tool for evaluating candidates for employment (Goldberg et al. 2006, Johnson 2014) and for mapping the psychological traits of individuals (Beck and Jackson 2022). For this reason, the current study adopts IPIP constructs as a tool for mapping key psychological traits of Operations and Supply Chain professionals, with the objective of identifying guidance concerning the relevant individuals' traits driving operational outcomes.

By adopting IPIP constructs and items, it has been possible to determine a taxonomy of psychological traits whose demand in relevant job descriptions can be measured by applying text mining techniques. This allowed us to obtain a profiling of OM and SCM professionals' noncognitive aspects by leveraging the IPIP framework and innovative big data analysis methodologies.

# 3 Methodology

Our methodology is based on the collection of relevant job descriptions in the OM and SCM domains, together with a relevant taxonomy of personality traits to be evaluated. Subsequently, we implemented a specific text mining process: we used the Semantic Brand Score (SBS) indicator (Fronzetti Colladon 2018) to evaluate the importance of specific terms in job vacancies, and the SBS BI app due to its wide range of capabilities in terms of text analysis (Fronzetti Colladon et al. 2021).

## 3.1 Job Description database selection

Job description data was collected and provided by Telpress International B.V. We provided specific research keywords to obtain relevant data related to OM and SCM professionals, as presented in Table 1.

**Table 1:** job description research keywords

| Operations keywords | Supply Chain keywords |
|---|---|
| Maintenance | Distribution |
| Manufacturing | Logistics |
| Operations | Supply Chain |
| Production | Transportation |

Keywords led to the identification of 5113 job descriptions published in the second half of 2023. However, 452 of them (8,8%) were not significant for the scope of our research (e.g., professionals in the marketing operations venue), leading to 4661 final job profiles. The

profiles were almost equally distributed among the two research areas (55,4% operations; 44,6% supply chain). Moreover, each job description had several attributes that allowed us to categorize the seniority level of the professional role (minimum: up to 3 years of experience; medium: up to 10 years of experience; high: more than 10 years of experience); the company size (small: up to 10 employees; medium: up to 50 employees; big: more than 50 employees); the company region (Europe; North America; and rest of the world). The classification of job descriptions is represented in Figure 2 and Figure 3.

**Figure 2: operations job descriptions clusterization**

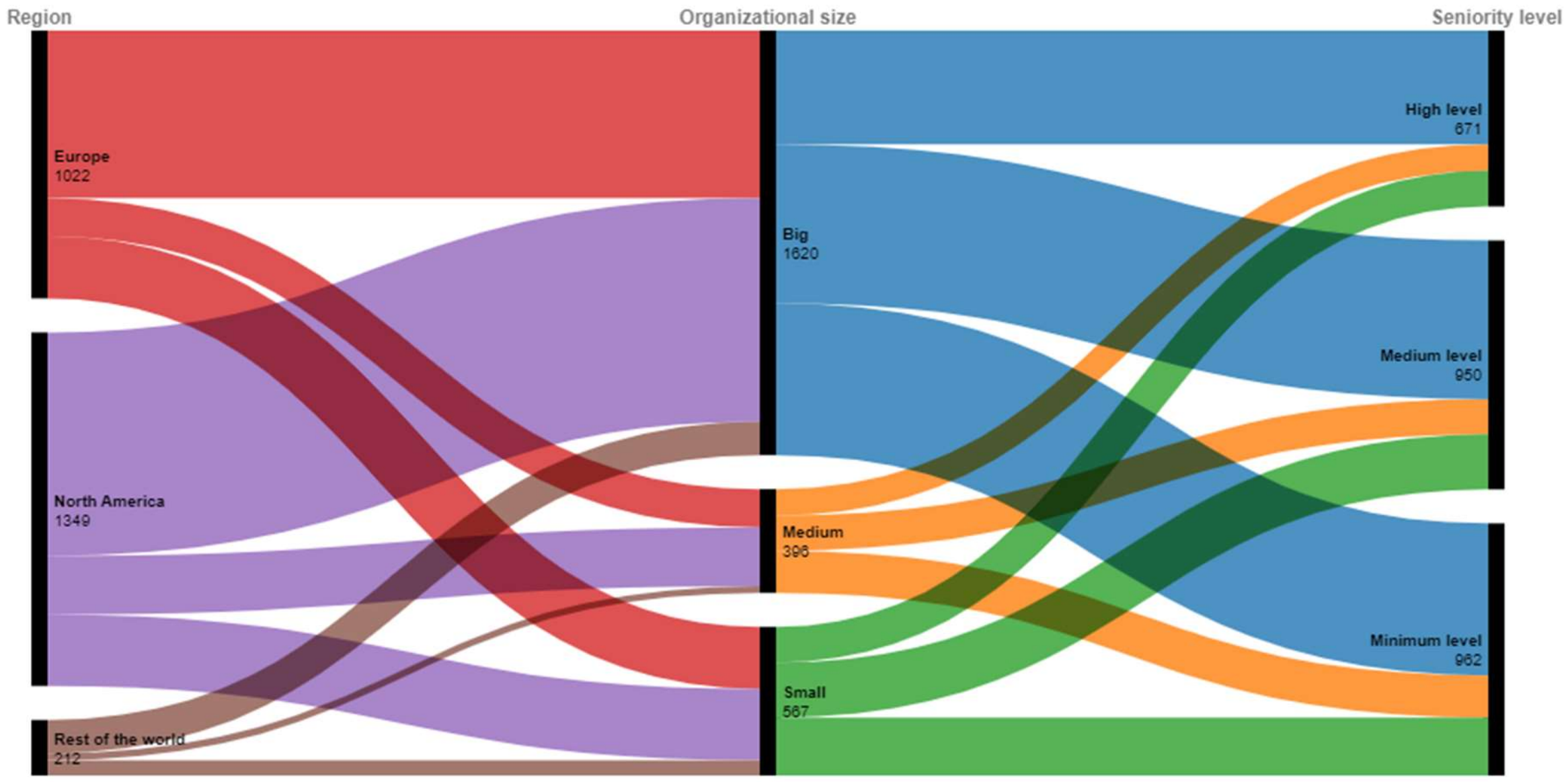

**Figure 3:** supply chain job descriptions clusterization

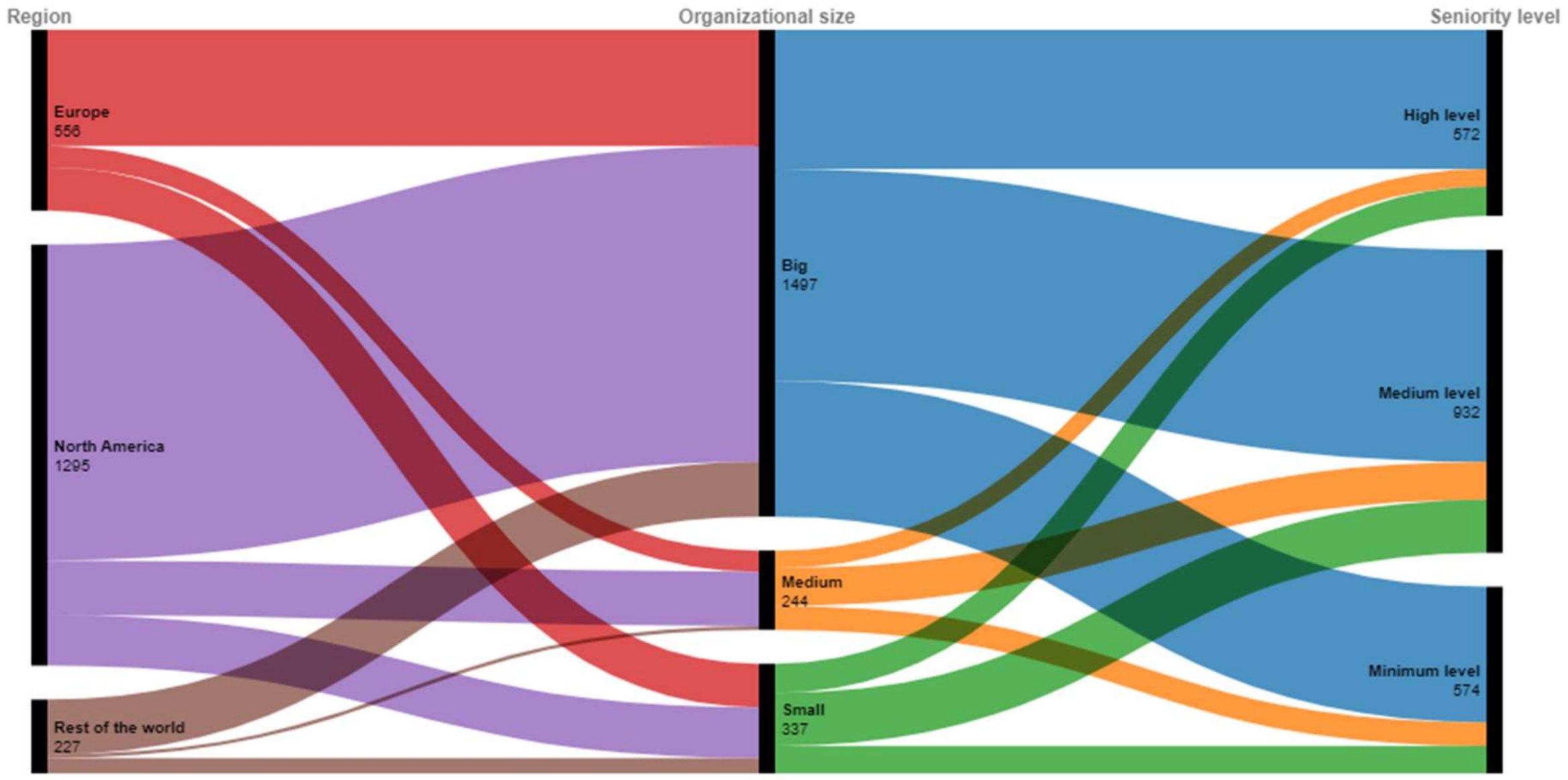

We can observe that the largest organizations in our dataset were mostly located in Europe and North America. It is also interesting to notice that the number of open positions in small companies was higher than in medium companies. The same considerations apply to organizations from the rest of the world, even though on a reduced scale.

## 3.2 Analyzing semantic importance

To analyze semantic importance and determine the demand for specific skills or personality traits in job descriptions, we used the Semantic Brand Score (SBS). This indicator leverages text mining techniques to evaluate the relevance and memorability of words in a given corpus of texts (Fronzetti Colladon 2018). The first step to compute the SBS was to employ text preprocessing routines – including tokenization, removal of stop-words, and word stemming (Porter 1980) – and to build semantic networks from job descriptions. These networks were then subjected to social network analysis for the SBS calculation by using the SBS BI web application (Fronzetti Colladon and Grippa 2020).

The SBS indicator consists of three dimensions: prevalence, diversity, and connectivity. The prevalence component measures the frequency of using a keyword in a text. A word – i.e., a skill – frequently appearing in job descriptions suggests that recruiters actively look for it in candidates. However, the importance of a keyword goes beyond its frequency of occurrence; it also depends on its associations with other words in the text. For example, when the term "capability" is linked to various other words, such as "leadership," "innovation," and "adaptability," it can significantly impact how individuals perceive the role of skills in career advancement. Conversely, if it is only associated with specific contexts like "technical capability" or "analytical capability," it can shape a more focused understanding of skill requirements in certain domains.

The SBS indicator also includes the dimensions of diversity and connectivity, which assess how varied and strong the associations to a skill-related keyword are and how effectively that concept can connect with other terms in the job description. The rationale is that the higher the frequency of a keyword within a job post and the richer its associations, the more impactful it will be.

Diversity, as part of the SBS index, evaluates how skill-related keywords are associated to other words in the text, drawing parallels to the construct of brand image (Keller 1993). When associations are less common and numerous, a keyword gains greater importance. Diversity is operationalized through a formula based on distinctiveness centrality (Fronzetti Colladon and Naldi 2020), which quantifies the significance of a keyword in relation to its neighbors in the network:

$$Diversity\ (i) = \sum_{\substack{j=1 \\ j \neq i}}^{n} \log_{10} \frac{(n-1)}{g_j} I(w_{ij} > 0)$$

Generally, our analysis involves a graph $G$, comprising $n$ nodes (representing words) and $E$ edges (representing word links), which are associated with a set of connection weights $W$. Words are linked when they appear close to one another in the text. Within the diversity formula, $g_j$ represents the degree of node $j$, which acts as one of the neighbors of node $i$ (the node for which diversity is calculated). We use the indicator function $I(w_{ij} > 0)$, which takes a value of $1$ when there is an edge connecting nodes *i* and *j* (i.e., when $w_{ij} > 0$) and a value of $0$ when this edge is absent.

The last dimension of the SBS index, connectivity, is measured as the weighted betweenness centrality of the skill-related keywords. It represents their 'brokerage power', indicating how effectively a keyword can act as a bridge, connecting other terms and topics in the discourse. The connectivity formula is based on the analysis of the shortest paths connecting pairs of nodes in the network:

$$Connectivity\ (i) = \sum_{j<k} \frac{d_{jk}(i)}{d_{jk}}$$

where $d_{jk}$ is the number of shortest network paths connecting nodes $j$ and $k$ (calculated using edge weights), and $d_{jk}(i)$ is the number of those paths that include node $i$.

The final SBS indicator is obtained by summing the standardized scores of its components, considering all words in the corpus for each time frame. Consistently with past research, an equal weighting scheme is applied, and standardization is performed by subtracting the mean and dividing by the standard deviation:

$$SBS\ (i) = \frac{PR_i - \overline{PR}}{std(PR)} + \frac{DI_i - \overline{DI}}{std(DI)} + \frac{CO_i - \overline{CO}}{std(CO)}$$

where $PR$ is prevalence, $DI$ is diversity, and $CO$ is connectivity.

Several studies have already implemented the SBS indicator since its first creation. To cite a few, the indicator has proven to be capable of forecasting election results by studying the importance of political candidates in online news (Fronzetti Colladon 2020) to predict museum visitors' variations through the analysis of online forums (Fronzetti Colladon et al. 2020) or to provide valuable insights for the improvement of energy communities social awareness (Piselli et al. 2022).

## 3.3 Lexicon selection

Normally, each company has the freedom to describe skills and attitudinal traits of job descriptions in its preferred manner, without any explicit categorization or reference to well-known frameworks. This made it challenging to determine the attribution between keywords and specific attitudinal traits reported by the IPIP database, also because psychological constructs do not have a unique way of expression.

IPIP is a widely used inventory that covers a comprehensive range of personality traits (Johnson 2014), whose adoption for big data analysis provided a well-established and extensively researched inventory, allowing access to a wide array of personality traits (Beck and Jackson 2022). This comprehensive coverage enhanced the accuracy and depth of insights extracted from text data, supporting more robust analyses and yielding a broader understanding of attitudinal nuances (Ringwald et al. 2023). From the IPIP constructs, we began the task of designing a vocabulary suited to our specific objective. The process involved carefully reviewing the selected psychological scale and extracting keywords that could encapsulate the essence of the corresponding construct. By carefully compiling this lexicon, we intended to capture the subtle attitudinal traits inherent in the job description.

The taxonomy we adopted to measure these constructs from text was built from a structured process that allowed the defining of the main relevant terms. We started the definition of the lexicon with a twofold approach. On the one hand, we analyzed the items corresponding to each psychological construct according to the IPIP database, deriving specific keywords. On the other hand, we performed a wide literature review to understand these constructs' inner meaning and nuances. This allowed us to obtain a lexicon for each scale.

Subsequently, we introduced a review process with a panel of 3 experts from relevant scientific and industrial fields. The selected experts reviewed the pre-identified lexicon and provided feedback on the chosen terms. The final lexicon selection was performed by revising the original keywords. Overall, the review process required a first plenary meeting and two additional meetings to discuss the cases where no agreement was reached yet. Note that this review allowed picking the most relevant keywords that could effectively convey the intended meaning of each attitudinal scale, considering their synonyms as well. In the appendix, we provide a table with examples of keywords for each construct.

Additionally, after selecting the keywords, we considered only the personality traits that emerged in at least 1% of the whole dataset of job descriptions as a significance threshold for our purposes. This stringent criterion ensured that only the most prominent and representative characteristics of the social spectrum were included in our subsequent analysis. By applying this selection process, we could streamline the initial pool of 244 original traits reported by IPIP to a final set of 28 traits. After selecting these constructs, a deeper analysis of their meaning and interpretation has been performed. We retrieved and studied the main scientific contributions discussing these traits. We provide a comprehensive definition for each psychological construct and interpret its possible values, which we then use for the subsequent interpretation of the SBS results. We provide these elements in Appendix B, specifically in Table B.1.

## 3.4 Topic modeling

We implemented topic modeling to examine key components and their relative significance in job descriptions. Using a network-based method (Fronzetti Colladon and Grippa, 2020; Lancichinetti et al. 2015), we identified topics by clustering the semantic network constructed from the job descriptions.

More specifically, topics were identified by applying community detection to cluster densely connected nodes in the semantic network (Blondel et al. 2008). The most relevant words in each cluster were selected based on weighted degree and the proportion of internal vs. external links. Words strongly connected within a cluster and loosely connected outside the cluster were considered descriptive of the topic. In summary, the network-based topic modeling approach looks at word co-occurrence patterns to identify topics, rather than using statistical modeling of word distributions. The word clusters comprise associated concepts that are discussed together frequently in the text. We adopted the SBS BI App for this purpose with the scope of completing the mapping of attitudinal traits and the evaluation of job descriptions (Fronzetti Colladon and Grippa 2020). To this aim, topic modeling of job posts was implemented at this stage to identify additional relevant keywords for the lexicon creation. Moreover, it was also adopted to determine the most common elements characterizing OM and SCM positions (Vayansky and Kumar 2020).

# 4 Results and discussion

## 4.1 Personality traits demand mapping of OM and SCM professionals

We started our analysis by exploring the main discourse themes emerging from job descriptions. The output is presented in Figure 4 for the OM job vacancies and Figure 5 for the SCM ones.

**Figure 4:** topic modeling for Operations Management job descriptions

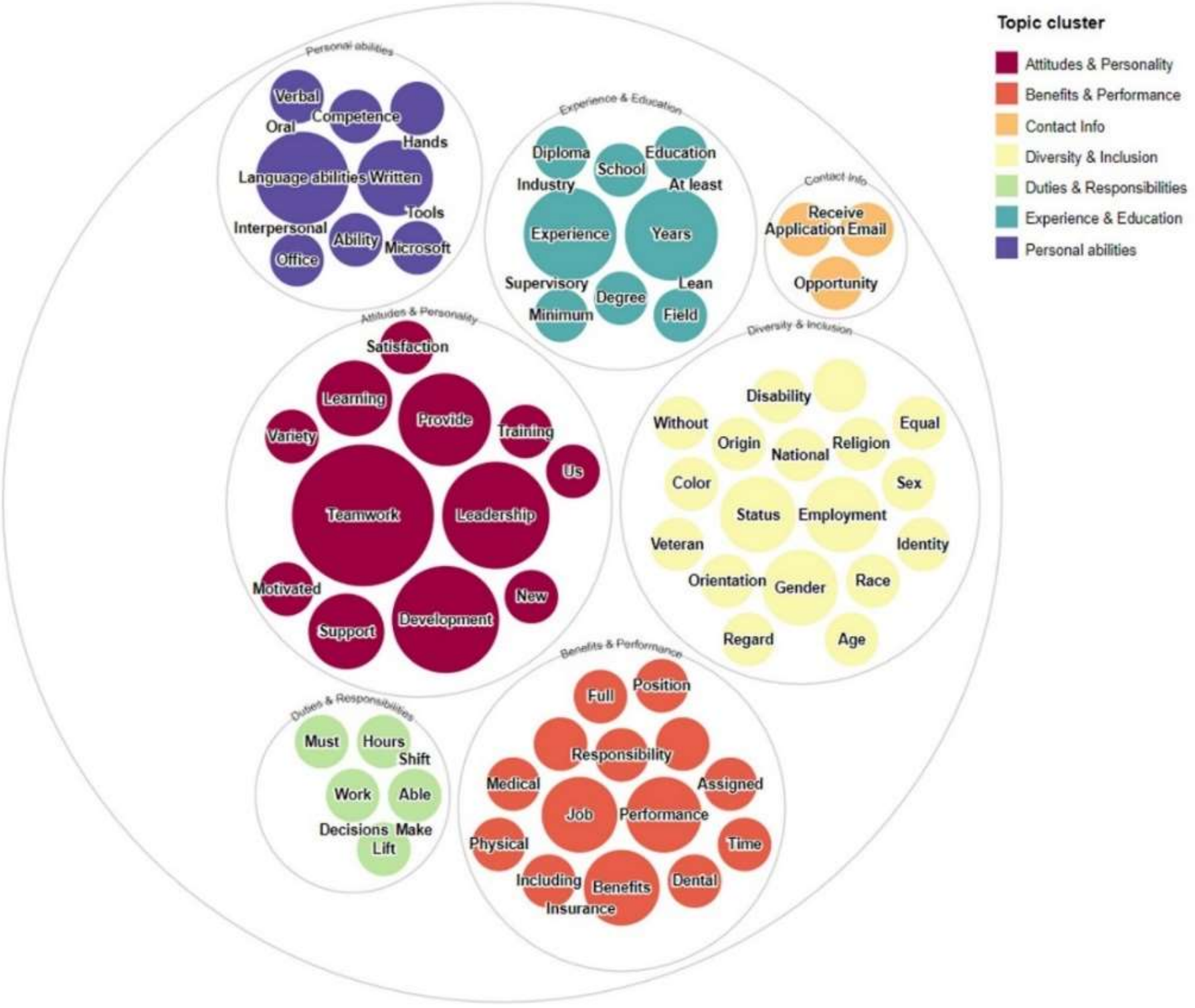

**Figure 5:** topic modeling for Supply Chain Management job descriptions

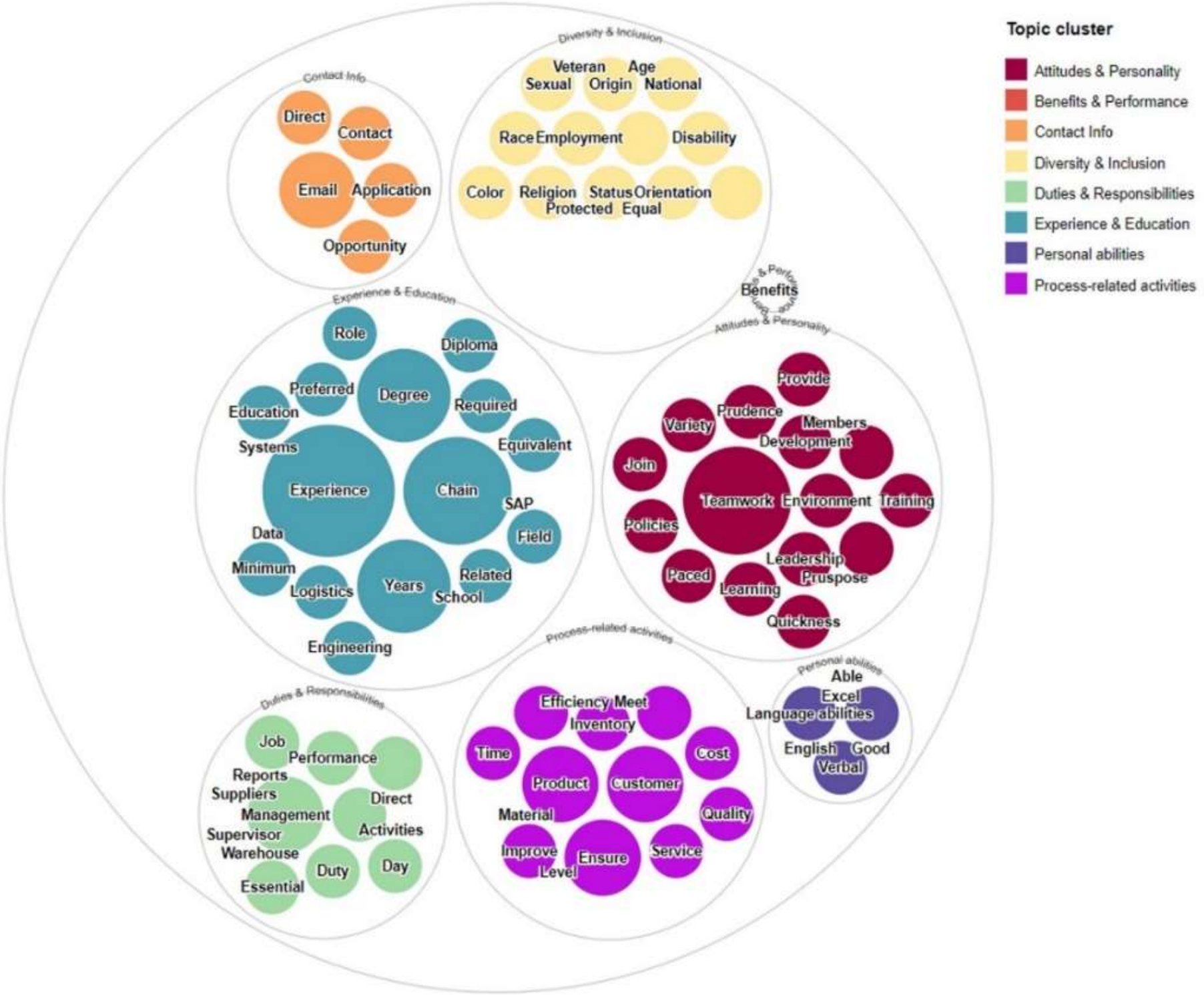

We can observe that the OM and SCM job profiles shared 7 topics. An additional topic emerged for SCM vacancies (*"Process-related activities"*). Note that the larger the circle, the greater the importance of the keyword and, in turn, of the topic cluster. A summary of the topic clusters is reported in Table 2, along with the relative weight of the cluster for the OM and SCM job descriptions.

**Table 2:** topic clusters description and relevance

| Topic cluster | Description | Relevance (OM) | Relevance (SCM) |
|---|---|---|---|
| Attitudes & Personality | Pertains to the desired attitudes and personality traits that an ideal candidate for the job position should have. | 18% | 19% |
| Benefits & Performance | Focuses on rewards and performance evaluation aspects. | 14% | 1% |
| Contact Info | Provides information regarding contact details. | 15% | 12% |
| Diversity & Inclusion | Addresses diversity and inclusion considerations. | 15% | 12% |
| Duties & Responsibilities | Describes the specific tasks and responsibilities. | 10% | 13% |
| Experience & Education | Emphasizes the required experience and educational background. | 13% | 21% |
| Personal Abilities | Highlights the necessary individual skills and abilities that an ideal candidate for the job position should have. | 14% | 7% |
| Process-related activities | Covers activities related to processes and procedures. | N/A | 15% |

From the analysis, we labeled the clusters and described them from their keywords. We obtained 8 clusters described as follows:

1. Attitudes & Personality: delves into the desirable attitudes and personality traits employers seek in an ideal candidate for the job position. The words in this cluster aim to paint a vivid picture of the kind of individual who would thrive in the role and fit well within the company culture. Employers may outline qualities such as strong work ethics, adaptability, teamwork, communication skills, problem-solving skills, and a positive attitude toward challenges. By emphasizing these traits, employers aim to attract candidates who not only possess the necessary skills but also possess the right mindset and character to succeed in the organization;

2. Benefits & Performance: focuses on two critical aspects for prospective candidates: the rewards they can expect and how their performance will be evaluated. With these terms, employers often detail the comprehensive benefits package offered to employees, including health insurance, retirement plans, paid time off, and other benefits. Additionally, the evaluation process of employee performance is highlighted, clarifying how success and progress are measured within the company. This information plays a vital role in motivating candidates to apply and demonstrating the company's commitment to rewarding and recognizing employees' efforts;
3. Contact Info: provides straightforward and practical details regarding how candidates can contact the hiring company. It typically includes the company's name, address, email address, and phone number. By clearly presenting contact information, employers aim to encourage potential applicants to initiate communication and inquire further about the job opportunity. This group of words facilitates a smooth and direct channel of communication between employers and candidates;
4. Diversity & Inclusion: addresses the company's commitment to fostering a diverse and inclusive work environment. Employers use this section to highlight their dedication to promoting diversity, equity, and inclusion within the organization. They may describe initiatives, programs, or policies aimed at creating an inclusive workplace where people from diverse backgrounds can thrive. The words used also reflect the company's values and show candidates that the company embraces diversity and values unique perspectives and experiences;
5. Duties & Responsibilities: outlines the specific tasks and responsibilities of the job position. Terms in this area clearly explain what the job entails and what is expected of the potential candidate. Employers list key job functions, roles, and deliverables, allowing candidates to gauge whether their skills align with the job requirements. By comprehensively detailing the duties and responsibilities, employers aim to attract applicants who possess the necessary expertise and experience to excel in the role;
6. Experience & Education: emphasizes the required experience and educational background that candidates should possess. Employers typically specify the minimum years of experience needed in a relevant field and outline the preferred educational qualifications, such as degrees or certifications. By specifying these requirements, employers aim to target candidates who have the necessary expertise and knowledge to contribute effectively to the organization;
7. Personal Abilities: highlights the specific individual skills and abilities an ideal candidate should have for the position. Employers may mention qualities and other essential attributes. These words provide candidates with information on the key competencies contributing to their success in the role. Employers aim to attract applicants who possess a unique set of talents that align with the job's demands;

8. Process-related Activities: covers various activities and procedures related to the job position. They may outline the steps involved in the application process, interview procedures, or any assessments or tests that candidates may need to undergo. Providing information on these activities helps candidates navigate the hiring process more effectively, making it transparent and efficient for both parties involved.

*"Attitudes & Personality"* represented the most prominent topic in OM job descriptions (accounting for 18% of overall weight), while it had the second position for SCM job descriptions (accounting for 19% of the overall weight) after *"Experience & Education"*. It was then followed by *"Diversity & Inclusion"* (respectively accounting for 15% and 12% of the overall weight of OM and SCM job descriptions). This outcome is comforting, as it highlights that organizations were paying increasing attention to human-related aspects during HR and hiring processes (Gonzalez et al. 2022, Kang et al. 2018), while great attention was placed on the background of applicants. On the other hand, relevant differences are found related to *"Benefits & Performance"* and *"Process-related activities"* aspects. Indeed, while OM job descriptions were much more developed regarding benefits and welfare statements, SCM ones were not. The contrary was true for the process description standpoint, whose elements are typically more important for SCM job profiles (note that the *"Process-related activities"* cluster was not present for the OM job profiles).

*As a conclusion of the initial topic modeling analysis, "Attitudes & Personality" has risen as one of the most prominent clusters. Therefore, it required a further deep dive that could be carried out with the SBS indicator. For this reason, the SBS evaluation was implemented as follows. The calculation of SBS values across the 28 IPIP constructs resulted in a detailed mapping derived from the analysis of our job descriptions, which is illustrated in Figure 6. The figure compares the calculated SBS values for the OM and SCM profiles and their absolute percent differences to distinguish between the two categories.*

**Figure 6:** psychological mapping of OM and SCM professionals

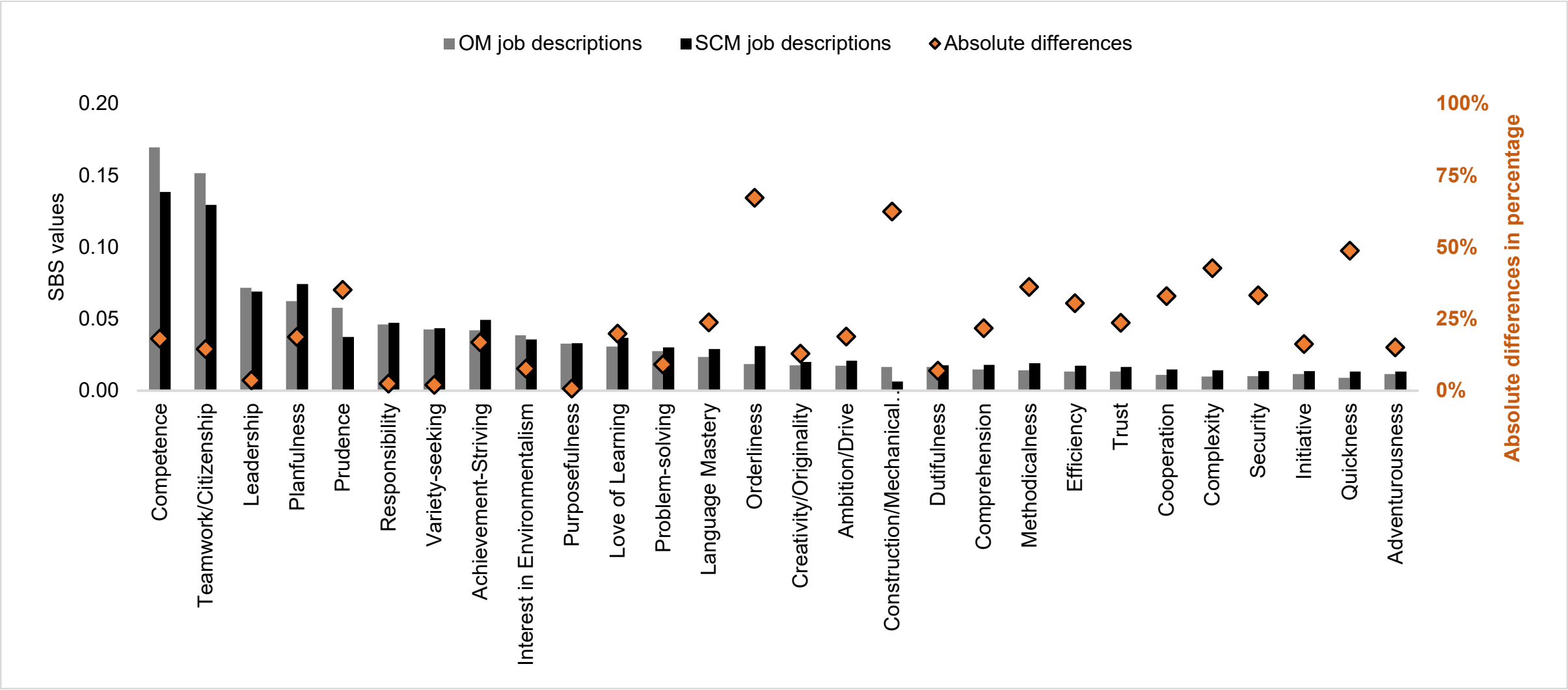

*"Competence"* – namely the "managerial potential" of an individual – and *"Teamwork/Citizenship"* – namely the orientation of an individual to sharing a group's obligations and duties – represented the most sought-after attitudinal characteristics. However, they were higher for Operations Management practitioners. This aligns with the results of Qin et al. (2015), where competence and cooperativeness are considered the foundation for achieving workforce agility in the OM context, and the research of Demirkan and Spohrer (2015), stating that the ability to collaborate, innovate, and demonstrate capabilities could effectively boost organizational performances (Van Oyen et al. 2001).

No major differences were encountered for *"Leadership"*, *"Responsibility"*, *"Variety-Seeking"*, *"Interest in Environmentalism"*, and *"Purposefulness"* traits, which were among the most relevant traits for the fields of OM and SCM. These traits have often been scrutinized by scientific research due to their perceived influence on organizational outcomes. For instance, Samson et al. (2018) adopted the multi-stakeholder decision theory to understand the employer-employee relationship when organizational and personal career outcomes misalignments occur. The authors observed that when a misalignment occurs, employees may make choices toward achieving personal gain. For this reason, having a long-term vision is crucial for executives and leaders who want to retain the top talents within their organization. This result is also influenced by the way of communicating that vision, as demonstrated by the exploratory work of Morreale et al. (2018). Following the same reasoning, Staats and Gino (2012) evaluated how the interplay between specialization and variety of tasks impacts operational performances, demonstrating that – even though task setup generates costs (Schultz et al. 2003) – they are increased by 23% with a varied task strategy over time, but with an intraday specialization. This result is also confirmed by the paper of Pisano et al.

(2016), which showed the importance of task variety for Operations and Supply Chain Management performances.

Furthermore, a substantial alignment between OM and SCM profiles was obtained for *"Problem Solving"*, *"Creativity/Originality"* and *"Initiative"* traits. While these traits are traditionally considered some of the most important attitudinal attributes of professionals in this venue (Dombrowski et al. 2017, Siebert and Kunz 2015), this was not confirmed by the SBS values, which were on the average-low end of the illustrated outcomes. A possible explanation can be given by taking into account organizational objectives and their vision. Since lateral thinking-related constructs refer to critical capabilities to ensure long-term organizational performance (Boudreau et al. 2003), companies may be currently focused on short/medium-term achievements that do not necessarily require those characteristics, which were hence reflected in the job descriptions.

Differently, major gaps were observed for *"Planfulness"*, *"Prudence"*, *"Achievement-Striving"*, among the traits with the highest SBS values. It should be noted that while *"Planfulness"* and *"Achievement-Striving"* values were more marked for SCM professionals – because they were considered the pre-requisite for success in SCM tasks (Boudreau et al. 2003, Stek and Schiele 2021) – the opposite held true for *"Prudence"*. An explanation of the latter result was given by the fact that SCM professionals are more exposed to risk-taking decisions – such as the definition of safety stock levels in uncertain conditions – which may lead to the development of an "uncertainty loving attitude" (Ancarani et al. 2013). On the contrary, OM professionals often show risk-averse behaviors, as shown by the research by Zhang et al. (2018).

Lastly, further notable considerations are given by looking at the *"Trust"* and *"Cooperation"* values, which surprisingly were at the very low end of the SBS values distribution. The role of trust has been widely analyzed in the current scientific literature, showing the importance of this trait, especially in supply chain decision-making processes. For example, reference can be made to the research work of Choi et al. (2020). Similarly, research (Allred et al. 2011, Hitt et al. 2016) described cooperation as a source of competitive advantage, with greater importance for SCM profiles than OM ones. Although the latter statement is consistent with the SBS values, those attitudinal traits do not appear to be as relevant as found by the scientific background. This is not to say definitively that these traits were not relevant for OM and SCM professionals, but it may suggest that they are difficult to assess in an interview or to specify in a job description. For instance, the sense of trust towards an employee is something built over time, often implicitly, and can only be evaluated after repeated interactions. This could be why recruiters and managers did not emphasize the construct in job vacancies.

## 4.2 Association of job features and psychological traits

A further refinement of the previous SBS computation was achieved by grouping the original datasets by job characteristics, namely, by considering the region of the organization issuing the job description, its organizational size, or the seniority level of the open position. In the following, we provide heatmaps comparing the SBS values for the 28 psychological constructs in terms of these 3 job features. The heatmaps are illustrated in Figure 8, and Figure 9.

### 4.2.1 Organization region

Data revealed a significantly higher emphasis on *"Teamwork/Citizenship"* among the required job profiles in Europe and the rest of the world compared to North America, with a more evident effect for Supply Chain Management professionals. This may suggest that professionals in Europe and the rest of the world value collaborative work environments and are inclined toward fostering synergistic relationships within teams with respect to a more individual and competitive work environment (Andreassi et al. 2014). Additionally, demand for Operations Management positions in Europe and the rest of the world exhibited a considerably higher required level of *"Competence" (managerial potential)* compared to their North American counterparts. This may imply that professionals in Europe and the rest of the world are perceived as having a stronger attitude toward managerial roles, yet this is not often well communicated on the outside, as shown by the research of Pollach and Kerbler (2011), even if this appears evident only for OM job profiles.

Interestingly, the data revealed that the demand for professionals in Europe and North America did not focus on *"Planfulness"* for both OM and SCM areas. This suggests that professionals outside of Europe and North America should have a higher propensity for structured and organized approaches in their work. This result may appear inconsistent with the current literature (Chang et al. 2017). However, it may be reasonable to think that the recruiters were looking for these characteristics, although they are not intrinsic to the specific culture. Surprisingly, a similar effect was observed for the *"Achievement-Striving"* attitudinal trait. North American job positions exhibited the lowest values among regions, suggesting a reduced focus on the drive to excel and achieve ambitious goals (which might be implicit in some cultures). Finally, the OM and SCM job positions in North America exhibited higher *"Prudence"* than their international counterparts. This surprising but notable finding indicates that Human Resources seeks professionals who prioritize cautious decision-making and risk management strategies.

**Figure 7:** SBS heatmap - Evaluation of regional variations in job positions

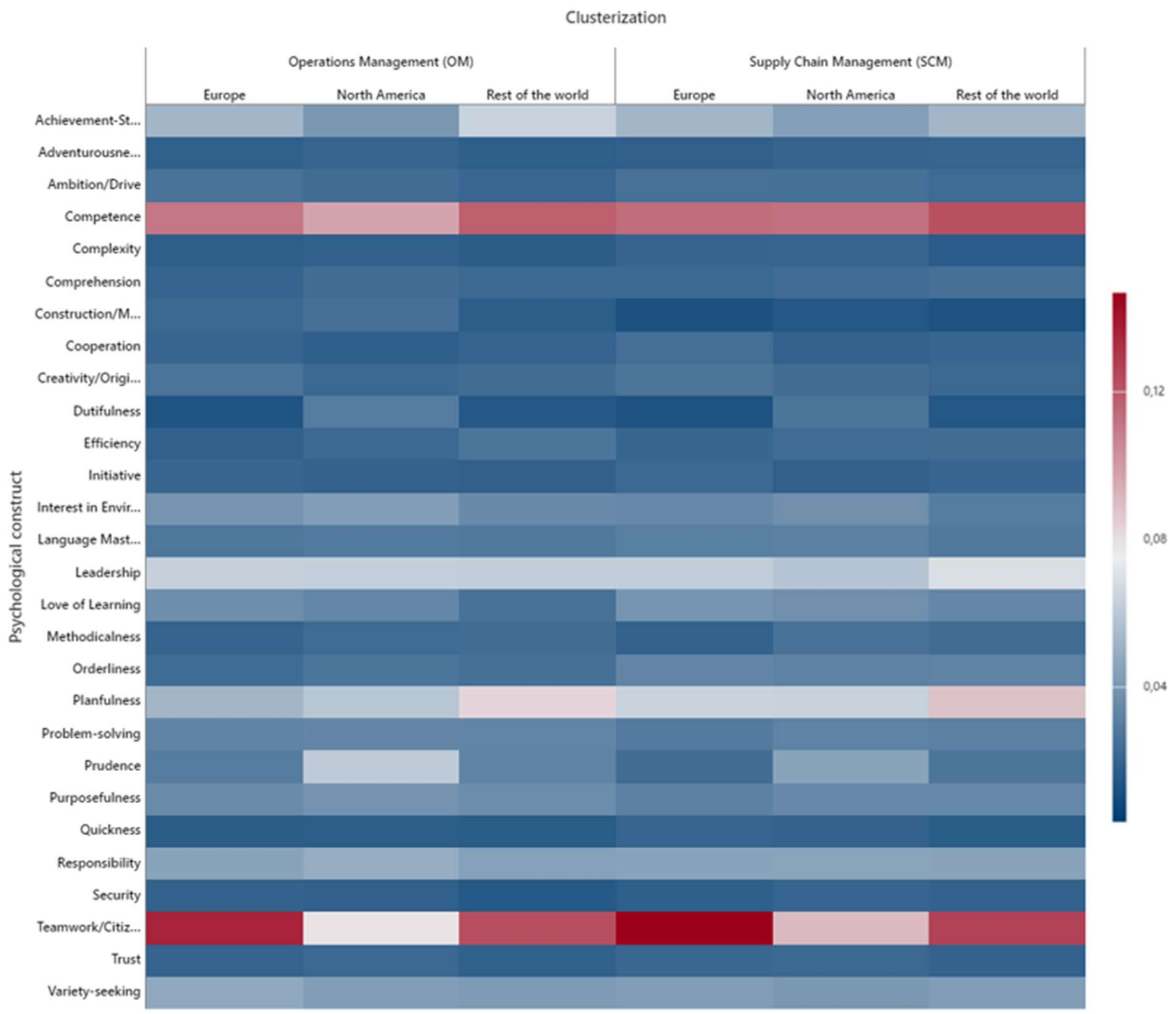

## 4.2.2 Organization size

Figure 8 indicates that for the *"Teamwork/Citizenship"* construct, OM job profiles in medium-sized organizations exhibit higher SBS values compared to those of large and small organizations. This could suggest that medium organizations place a greater emphasis on collaboration and cooperation (Cassell et al. 2002). These findings could be attributed to the complexity of tasks in medium-sized organizations, which are less structured than those in larger organizations' functional areas. Notably, a different outcome is achieved by the *"Competence"* trait. Indeed, job profiles for large organizations show a higher SBS value, which is even higher for Operations Management rather than Supply Chain Management. This observation could be attributed to the larger scale and scope of operations within large organizations, which necessitate a higher level of managerial expertise and strategic decision-making.

*"Planfulness"* was found to be higher in job profiles of small organizations. This suggests that small organizations prioritize meticulous planning and organizational skills in their OM and SCM professionals, potentially due to the need to optimize resources and overcome limitations associated with limited budgets and capacity (Curado 2018). Similarly, requests for professionals in small organizations exhibited a higher sense of *"Responsibility"* than their requirements for counterparts in medium and large companies, highlighting the importance placed on individual accountability and ownership of tasks within small organizations. This effect may stem from the need for employees to fulfill multiple roles and manage diverse responsibilities with respect to larger organizations with more standardized tasks (Krishnan and Scullion 2017). Lastly, we found that the requirements for professionals in large organizations display a slightly higher inclination toward "*Variety-Seeking"* in their job positions, showing the need for adaptability and versatility that can contribute to this trait's prominence.

**Figure 8:** SBS heatmap - Evaluation of organizational-size-related variations in job positions

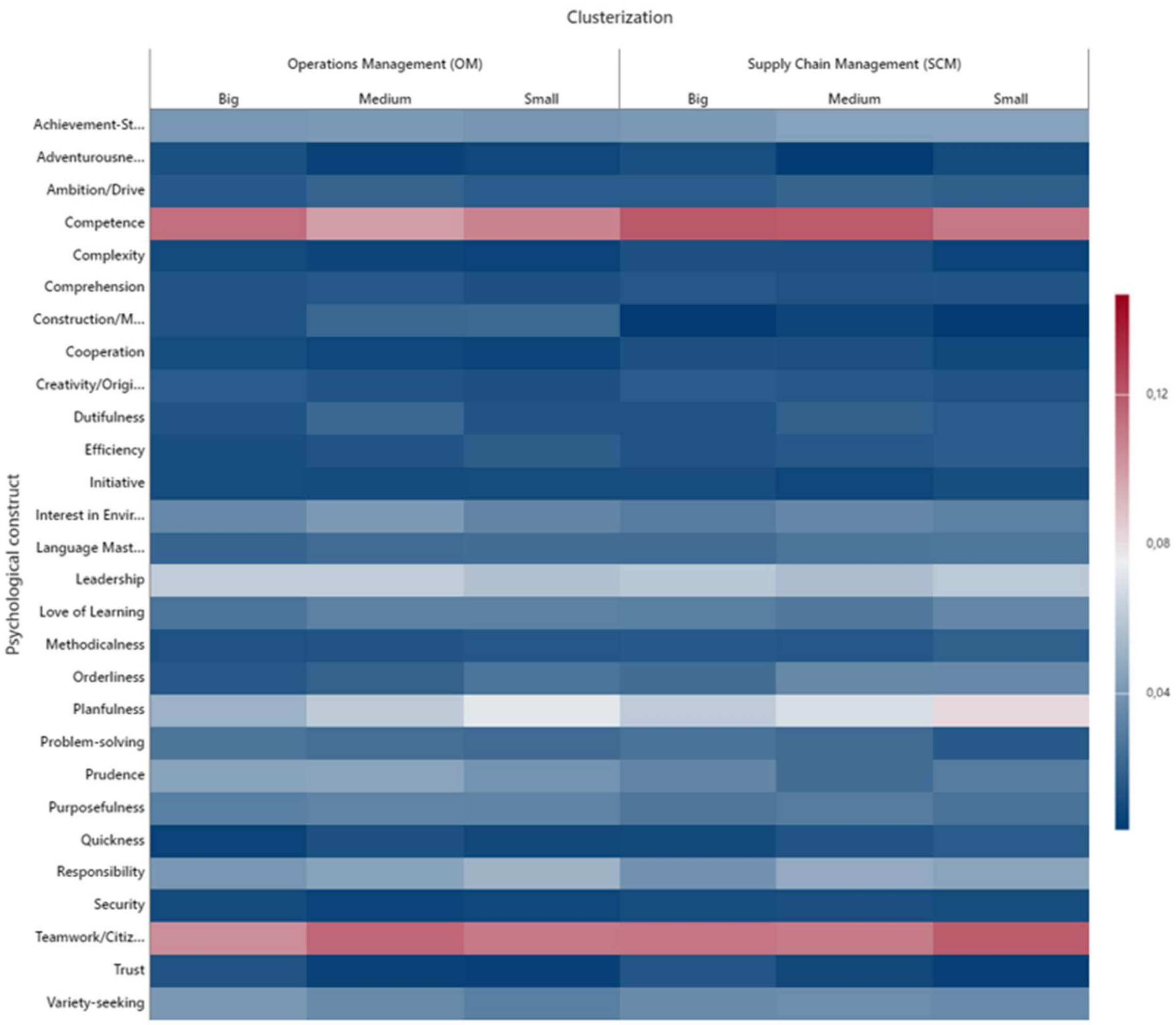

### 4.2.3 Seniority level of the position

As Figure 9 shows, *"Teamwork/Citizenship"* was found to be a highly required attitude for executive, top, and middle managers concerning entry positions. This was even more pronounced for Supply Chain Management professionals. This result suggests that senior-level professionals in OM and SCM are expected to possess strong collaborative skills (Andreassi et al. 2014) and exhibit a high propensity for teamwork. Such emphasis on teamwork at higher levels probably stems from the complexity and interdependence of tasks involved in managerial roles, where coordination and cooperation are of great importance. A similar result was observed for *"Leadership"* and *"Planfulness"*. Indeed, these traits were considerably higher for high- and medium-level positions compared to entry-level ones. This result aligns with previous research, as managerial roles require strong leadership skills to guide and inspire teams, make critical decisions, and foster a culture of excellence within the organization (Davidson-Schmich et al. 2023). The emphasis on leadership as seniority increases suggests that OM and SCM professionals at higher levels are expected to assume greater responsibilities and exhibit long-term perspectives and visionary qualities. Following the same considerations, the finding for *"Planfulness"* highlights that individuals in executive and managerial roles must possess a strategic mindset, enabling them to develop and implement long-term plans.

However, contrary to initial expectations, our findings indicate that entry-level positions exhibit higher levels of required *"Competence"* than higher-level positions. This unexpected outcome may be attributed to the need for individuals in entry-level roles to demonstrate their attitude and potential for growth, reflecting a focus on identifying future leaders within the organization (Krishnan and Scullion 2017). However, entry-level people are expected to possess more technical skills than broader managerial skills due to the more operative tasks they have to perform.

Interestingly, the SBS values for *"Variety-Seeking"* showed the same tendencies, being higher for entry-level positions. This outcome suggests that individuals in entry-level roles should demonstrate a propensity to explore diverse tasks and responsibilities to grow professionally. Such a required inclination for variety-seeking may stem from a desire to acquire a broad skill set, gain exposure to different facets of the profession, and discover areas of interest for future career development.

**Figure 9:** SBS heatmap – Evaluation of seniority level-related variations in job positions

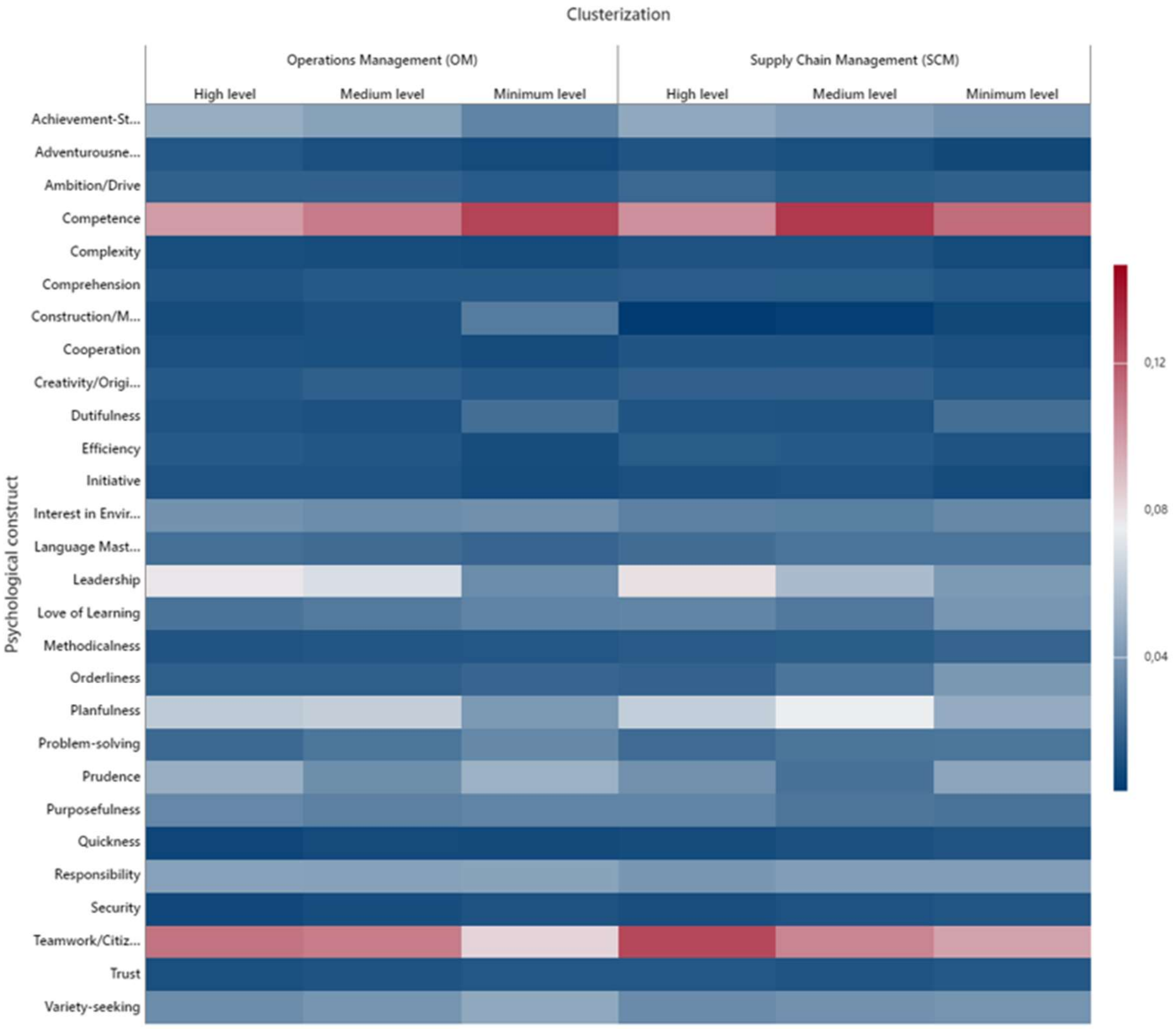

# 5 Conclusions

Our study proposes an innovative methodology based on semantic network analysis techniques for the psychological mapping of Operations Management (OM) and Supply Chain Management (SCM) characteristics from job descriptions. The methodology was applied to a set of 4661 job descriptions published in the second half of 2023 and allowed to derive a comprehensive psychological mapping of required job profiles in the OM and SCM domains by calculating the Semantic Brand Score measurement of significant attitudinal constructs.

Our analysis demonstrates that contextual factors and specific job characteristics, such as organizational region, size, and seniority level, influence the demand for attitudinal requirements, highlighting the need for nuanced talent management and recruiting strategies. In summary, *"Competence"* and *"Teamwork/Citizenship"* were identified as the most requested non-cognitive aspects for professionals in OM and SCM venues. However, differences emerged when the data was broken down by region or seniority level. These aspects were emphasized more in Europe and the rest of the world than in North America, particularly in job positions from medium and large organizations. Regarding the seniority of positions, *"Competence"* was more highly valued for entry-level roles than senior positions. Distinctive patterns also emerged for other traits like *"Leadership"*, *"Responsibility"*, *"Variety-Seeking"*, *"Interest in Environmentalism"*, *"Purposefulness", Planfulness"*, *"Prudence"* and *"Achievement-Striving"*. Contrary to expectations, traits such as *"Problem Solving"*, *"Creativity/Originality"*, *"Initiative", "Trust"* and *"Cooperation"* were less demanded, despite traditionally being considered some of the most relevant characteristics of an OM and SCM professional.

The findings of this study have several implications for human resources managers and the development of professional initiatives within the OM and SCM domains. By understanding the psychological traits most valued for specific job roles, as revealed by the study, organizations can more accurately define the attributes they seek from potential employees. This alignment between job requirements and desired psychological characteristics can help organizations streamline their recruitment and selection procedures. Furthermore, insights from our analysis can inform the design of targeted training and development programs to enhance the effectiveness and success of professionals in these roles. This latter observation is particularly relevant to answer the growing need for personalized learning pathways.

Our work contributes to the Human Resource Management literature and to research in Operations and Supply Chain Management. It introduces a research framework and a specific methodology that allows for determining the most requested skills and personality traits from the job market by leveraging text mining techniques and an innovative indicator (the Semantic Brand Score). The method is useful for scholars as it provides a systematic and comprehensive way to analyze job descriptions, efficiently extracting meaningful insights from textual data. It should also be noted that our tools can be useful for continuously monitoring companies' job vacancies and those of competitors, that is, to determine which skills are needed in a specific market and how to optimize a company's job postings.

Building on the current work, future research could explore the longitudinal impact of the identified psychological characteristics on operational performance and organizational outcomes. This could involve investigating the relationships between attitudinal traits, job satisfaction, employee engagement, and performance metrics to understand their interplay better. Furthermore, extending the analysis to include a wider range of industries and job

positions – not only limited to the OM and SCM industrial domains – would provide valuable insights into attitudinal differences across diverse operational contexts.

In conclusion, our research contributes to advancing behavioral operations research by also providing an understanding of the attitudinal traits that differentiate operations and supply chain professionals. To the best of our knowledge, by employing an innovative methodology and considering contextual factors, this research is the first to comprehensively map the demand for psychological traits characterizing OM and SCM professionals.

# Appendix A: publications in behavioral operations research

The following table reports the 58 publications in the behavioral operations research venue we considered for this study.

**Table A.1:** publications in behavioral operations research

| Reference | Contribution type | Journal/Publisher |
|---|---|---|
| (Schweitzer and Cachon 2000) | Paper | Management Science |
| (Van Oyen et al. 2001) | Paper | IIE Transactions |
| (Ahmad and Schroeder 2003) | Paper | Journal of Operations Management |
| (Boudreau et al. 2003) | Paper | Manufacturing & Service Operations Management |
| (Schultz et al. 2003) | Paper | Journal of Operations Management |
| (Loch and Wu 2005) | Book | Foundations and Trends in Technology, Information, and Operations Management |
| (Bendoly et al. 2006) | Paper | Journal of Operations Management |
| (Gino and Pisano 2008) | Paper | Manufacturing & Service Operations Management |
| (Loch and Wu 2008) | Paper | Management Science |
| (Bendoly et al. 2010) | Paper | Production and Operations Management |
| (Ho et al. 2010) | Paper | Management Science |
| (Tokar 2010) | Paper | The International Journal of Logistics Management |
| (Allred et al. 2011) | Paper | Decision Sciences |
| (Kalkanci et al. 2011) | Paper | Management Science |
| (Katok and Siemsen 2011) | Paper | Management Science |
| (Nair et al. 2011) | Paper | Journal of Operations Management |
| (Staats and Gino 2012) | Paper | Management Science |
| (Ancarani et al. 2013) | Paper | International Journal of Production Economics |
| (Croson et al. 2013) | Editorial | Journal of Operations Management |
| (Lee et al. 2013) | Paper | International Journal of Production Economics |
| (Muduli et al. 2013) | Paper | Resources, Conservation and Recycling |
| (Ren and Croson 2013) | Paper | Management Science |
| (Strohhecker and Größler 2013) | Paper | International Journal of Production Economics |
| (Wu 2013) | Paper | International Journal of Production Economics |
| (Zhang and Chen 2013) | Paper | International Journal of Production Economics |
| (Zhao et al. 2013) | Editorial | International Journal of Production Economics |
| (Long and Nasiry 2014) | Paper | Management Science |
| (Demirkan and Spohrer 2015) | Paper | Research-Technology Management |
| (Mura et al. 2015) | Paper | International Journal of Operations & Production Management |
| (Qin and Nembhard 2015) | Paper | Surveys in Operations Research and Management Science |
| (Siebert and Kunz 2015) | Paper | European Journal of Operational Research |
| (Sood and Sharma 2015) | Conference paper | Procedia - Social and Behavioral Sciences |
| (Franco and Hämäläinen 2016) | Editorial | European Journal of Operational Research |
| (Hitt et al. 2016) | Paper | Journal of Operations Management |
| (Pisano et al. 2016) | Case Study | Harvard Business School |
| (Tong and Feiler 2016) | Paper | Management Science |
| (Melnyk et al. 2017) | Paper | International Journal of Production Research |
| (Witt et al. 2017) | Paper | International Journal of Services and Operations Management |
| (Huang et al. 2018) | Editorial | Annals of Operations Research |
| (Kotzab et al. 2018) | Paper | Supply Chain Management: An International Journal |
| (Liu et al. 2018) | Paper | Annals of Operations Research |
| (Morreale et al. 2018) | Paper | Annals of Operations Research |
| (Samson et al. 2018) | Paper | Annals of Operations Research |
| (Shipley et al. 2018) | Paper | Annals of Operations Research |
| (Zhang et al. 2018) | Paper | Annals of Operations Research |
| (Zhao et al. 2018) | Paper | Annals of Operations Research |
| (Donohue et al. 2019) | Book | Wiley |
| (Erjavec et al. 2019) | Paper | Economic research |
| (Liu et al. 2019) | Paper | Modern Supply Chain Research and Applications |
| (Timmer and Kaufmann 2019) | Paper | Journal of Supply Chain Management |
| (Choi et al. 2020) | Paper | Management Science |
| (Donohue et al. 2020) | Paper | Manufacturing & Service Operations Management |
| (Erjavec and Trkman 2020) | Paper | International Journal of Services and Operations Management |
| (Romero and Stahre 2020) | Conference paper | 54th CIRP Conference on Manufacturing Systems |
| (Aloini et al. 2021) | Paper | TQM Journal |
| (Croom et al. 2021) | Paper | Journal of Purchasing and Supply Management |
| (Katok and Villa 2021) | Paper | Management Science |
| (Stek and Schiele 2021) | Paper | Journal of Purchasing and Supply Management |

# Appendix B: definitions of psychological constructs and value interpretations

The following table describes the 28 psychological constructs included in our analysis. We provide a brief definition, a description of value interpretations, and some references related to each dimension.

**Table B.1:** psychological construct definitions and value interpretation

| Dimension label | Dimension definition | Value interpretation | References |
|---|---|---|---|
| Achievement-Striving | The will of an individual to strive for excellence, reflecting the desire to obtain success and the individual's aspirations. | Individuals scoring high on this construct show great desire toward achieving successful results and high aspirations, although too high scorers may show workaholic behavior.<br><br>Individuals scoring low on this construct show an absence of ambitions and may manifest lazy behaviors. | (Costa et al. 1991, Hill et al. 2012, Johnson 2014, Moon 2001) |
| Adventurousness | The orientation of an individual to continuously look for adventures and be excited when facing the unknown. | Individuals scoring high on this construct manifest behaviors directed to searching for adventures and strong feelings, even though they may be dangerous or unknown.<br><br>Individuals scoring low on this construct typically prefer to experience ordinary situations, avoiding uncertainty or potentially dangerous situations. | (Cheung et al. 2003, Johnson 2014, Proctor and McCord 2009) |
| Ambition/Drive | The tendency to act according to one's ambitions and desires, which a dependence on individual and social rewards may drive. | Individuals who score high on this construct typically aim to obtain rewards for their actions and fulfill their strong ambitions. Very high scorers may manifest anger or addiction-related behaviors if the desired rewards are not achieved.<br><br>Individuals scoring low on this construct do not show strong ambitions and are not driven by rewards; rather, they usually adopt self-centered behaviors. | (Carver and White 1994, Cooper et al. 2008, Costumero et al. 2016, Franken 2002) |
| Competence | The ability of an individual to show capability and reliability to access new ideas and understand others' behavior. This construct also reflects the "managerial potential" of an individual. | Individuals scoring high on this construct are able to transmit a sense of competence, capability, and understanding to other people, hence being very reliable and open-minded. Very high scorers may show proactive behaviors toward innovation.<br><br>Individuals who score low on this construct show a lack of ability to present themselves as reliable and competent. Low scores may indicate a negative predisposition to managerial potential. | (Cartwright and Wink 1994, Costa et al. 1991, Hogan, Hogan and Murtha 1992, Rai et al. 2015) |
| Complexity | The breadth of outlook of an individual, being open to new ideas, experiments, and ways of doing/seeing things. | Individuals scoring high on this construct typically have critical and flexible minds, hence being experimental and open to change. Very high scores may indicate a liberal nature of the person.<br><br>Individuals scoring low on this construct show cautious behaviors and traditional rather than experimental approaches to different situations. Very low scores may indicate a conservative nature of the person. | (Agronick and Duncan 1998, Mitchell 1963, Zhang and Chen 2013) |
| Comprehension | The comfort of an individual with intellectual and sophisticated thinking. This construct is usually associated with learning, reading, and speaking abilities. | Individuals scoring high on this construct are typically predisposed to complex and conceptual discussions.<br><br>Individuals scoring low on this construct are usually not inclined toward complex thinking, facing discomfort when exposed to conceptual discussions. | (Plax and Rosenfeld 1979, Tutton 1996, Woods et al. 2019) |
| Construction/ Mechanical Interests | The orientation to have interests in construction and/or mechanical activities. This construct is mainly related to the predisposition to practical and manual tasks. | Individuals scoring high on this construct are typically inclined to carry out practical tasks and are interested in mechanical/construction speculations rather than conceptual ones.<br><br>Individuals scoring low on this construct do not show a predisposition to mechanical and practical interests, being more inclined toward conceptual investigations. | (Pozzebon 2012, Pozzebon et al. 2010) |
| Cooperation | An individual's tendency toward social cohesion and to support and actively collaborate with others. | Individuals who score high on this construct typically establish fruitful and stable relationships with others, generating constructive discussions and successful collaborations.<br><br>Individuals scoring low on this construct show an orientation toward acting by their personal perspective, typically imposing their own will on others. | (Johnson 2014, Mitchelson et al. 2009, Ross et al. 2003) |
| Creativity/Originality | The capability of an individual to generate innovative and unusual ideas, ways of acting and thinking. | Individuals scoring high on this construct are inclined to identify adaptive and novel ideas, which may represent a breakthrough in the individuals' common ways of acting and thinking. Very high scores for this construct may indicate eccentric behavior.<br><br>Individuals who score low on this construct typically cannot generate innovative and unusual ideas. | (Driskell et al. 1994, Johnson 1994, Oluf and Furnham 2015, Peterson and Seligman 2004) |
| Dutifulness | The observation and adherence of an individual to standards of conducts, ethical principles, and moral obligations. Note that it does not consider the origination and abstraction of moral principles but only the individual's adhesion to them. | Individuals scoring high on this construct typically act by conscience and are considered to "do the right thing," not only for themselves but also for the community.<br><br>Individuals scoring low on this construct tend to act in their interests, failing to adhere to moral principles and obligations. | (Costa et al. 1991, Johnson 2014, Moon 2001) |
| Efficiency | The ability of an individual to be diligent in performing activities and properly start and follow through with scheduled tasks. | Individuals scoring high on this construct typically act without a push and can effectively and properly use their time. Very high scorers may manifest excessively rigid behaviors.<br><br>Individuals scoring low on this construct usually adopt a chaotic approach to activities, with a tendency to forget things and inefficiently use their time. | (Bucher and Samuel 2019, Jackson et al. 2010, Lodi-Smith et al. 2010) |
| Initiative | The orientation toward being industrious and hardworking, showing proactive behaviors and individual initiative when carrying out activities. | Individuals who score high on this construct typically show enterprising and energetic behavior when carrying out tasks. Very high scorers may manifest innovation-related behaviors.<br><br>Individuals scoring low on this construct usually rely on others' initiative and ideas to get on with things. Very low scorers may show traits associated with laziness. | (Martinotti et al. 2008, Piskunowicz et al. 2014, Zhang and Chen 2013) |
| Interest in Environmentalism | The tendency toward having habits and attention to preserve and reduce the impacts of human activities on the environment. | Individuals scoring high on this construct typically have green habits with reduced environmental impact and aim at preserving nature from human impact. Very high scores may be associated with behaviors meant to raise other citizens' environmental awareness.<br><br>Individuals scoring low on this construct adopt short-term thinking and usually exhibit behaviors that could harm the environment in the long run. | (Goldberg 2010, Skimina et al. 2019) |
| Language Mastery | The interest of an individual toward scholarly activities and the inclination toward developing a comprehensive language. | Individuals who score high on this construct typically have a predisposition to academic achievement and the development of a rich vocabulary. These individuals can use their language and knowledge to make others understand their thinking and ideas.<br><br>Individuals scoring low on this construct do not show inclinations toward scholarly activities and typically do not master the language usage, finding difficulties in making others understand their ideas. | (Hogan, Hogan and Gregory 1992, Hogan and Hogan 2007, Pozzebon 2012, Pozzebon et al. 2010) |
| Leadership | The orientation and ability to lead, influence, and motivate others' actions toward establishing good relationships and collective success. | Individuals who score high on this construct typically show traits useful to hold a leadership role. They can lead, motivate, and influence others.<br><br>Individuals who score low on this construct lack the attributes to lead and motivate others effectively. | (Peterson and Seligman 2004, Pozzebon 2012, Schwaba et al. 2020) |
| Love of Learning | The inclination of an individual to continue learning and satisfy their curiosity by increasing their level of knowledge. This construct is associated with a positive feeling when learning new skills. | Individuals who score high on this construct typically perform activities geared toward continuous learning and expanding their knowledge and skills, such as reading books or studying different topics.<br><br>Individuals scoring low on this construct do not manifest interest in expanding their knowledge and skills, seldom being interested in only a few specific subjects. | (Goldberg 2010, Peterson and Seligman 2004) |
| Methodicalness | The orientation and need of an individual to observe a cognitive and ordered structure when approaching activities. | Individuals scoring high on this construct typically adopt structured methodologies and logical reasoning to perform activities, hence using specific cognitive structures. Very high scorers may manifest traits associated with low impulsivity.<br><br>Individuals who score low on this construct usually do not adopt logical mental structures when performing activities. Low scorers may manifest high impulsivity. | (Jackson et al. 1996, Loehlin 2012) |
| Orderliness | The inclination toward careful, conscientious, and meticulous behaviors that define highly organized individuals. | Individuals scoring high on this construct tend to be satisfied and look for order and neatness, being cautious and meticulous. Very high scorers may like to live in a routinary way.<br><br>Individuals scoring low on this construct usually have unsystematic behavior and live in an untidy manner. | (Caprara et al. 1995, Hahn and Comrey 1994, Johnson 2014) |
| Planfulness | The orientation toward planning, self-discipline, and fitting in structured settings. | Individuals scoring high on this construct typically have the predisposition to live in structured and highly defined settings, with an inclination to activity planning and scheduling.<br><br>Individuals scoring low on this construct do not fit in highly structured environments. Very low scorers may feel creatively constrained by structured and schematic settings. | (Cartwright and Wink 1994, Detrick and Roberts 2021, Domino 1971) |
| Problem-solving | The tendency of an individual to see "out of the box" and effectively solve and manage complex problems. | Individuals scoring high on this construct can identify innovative solutions to solve complex problems, showing analytical and out-of-the-box approaches. Very high scorers can typically act under pressure without negatively impacting their performance.<br><br>Individuals scoring low on this construct usually do not adopt analytical thinking and cannot generate out-of-the-box solutions. Very low scorers may suffer from acting under pressure with a deterioration of their overall performance. | (Assessment 2021, Hogan and Hogan 2007) |
| Prudence | The orientation toward carefully evaluating choices and adopting long-term thinking, to preserve one's future. | Individuals scoring high on this construct show behaviors that lean toward avoiding unnecessary risks and eventually sacrificing short-term pleasures to achieve more satisfying long-term objectives.<br><br>Individuals scoring low on this construct manifest behaviors oriented toward short-term pleasures, which usually sacrifice long-term goals and may cause regrets in one's future. | (Lee and Ashton 2018, Peterson and Seligman 2004, de Vries et al. 2020) |
| Purposefulness | The ability to have a clear purpose and manifest personal intentions. | Individuals who score high on this construct are generally perceived as confident and purposeful, showing a clear direction with their actions and reasoning. Very high scorers may also be perceived as calm individuals.<br><br>Individuals scoring low on this construct typically take actions without a clear direction, being perceived as disoriented. Very low scorers may also be perceived as incompetent. | (Hofstee et al. 1992, Lodi-Smith et al. 2010, Roberts et al. 2005) |
| Quickness | The inclination to easily understand things and handle massive amounts of information. | Individuals who score high on this construct can manage complex situations and information with an analytical approach.<br><br>Individuals who score low on this construct generally manifest difficulties in easily understanding complex situations and performing analytical reasoning with a large amount of data to handle. | (Driskell et al. 1994, Hofstee et al. 1992, Schwaba et al. 2020) |
| Responsibility | The orientation toward having integer and honest behaviors when relating to others and being accountable for individual actions. | Individuals scoring high on this construct typically feel strongly obligated to be integer and honest to others and the community. Very high scorers typically have a rigid conscience.<br><br>Individuals scoring low on this construct tend to be self-centered, easily break promises, and have a flexible conscience oriented toward self-interest. Very low scorers may refuse to account for their actions. | (Jackson 1994, Paunonen and Jackson 1996, Soto and John 2017) |
| Security | The tendency of an individual to feel stable, calm, and self-comfortable. | Individuals scoring high on this construct typically show an imperturbable mindset; they are comfortable with themselves and their actions. Very high scorers may be perceived as capable individuals.<br><br>Individuals who score low on this construct typically manifest unstable behaviors and discomfort with themselves. | (Detrick and Roberts 2021, Gough 1969, Hogan 1989, Hogan et al. 1984) |
| Teamwork/ Citizenship | An individual's orientation to identify with a group, share the group's obligations and duties, and be loyal to their fellow members. | Individuals scoring high on this construct tend to identify themselves with the activities and duties of a group, thus effectively supporting its cause.<br><br>Individuals scoring low on this construct manifest self-oriented behaviors, being more interested in their own cause than the community's. | (Peterson and Seligman 2004, du Plessis and de Bruin 2015) |
| Trust | The lean toward holding beliefs concerning the good intentions and behaviors of others. It is opposed to "distrust", which is the tendency to hold negative beliefs and suspicions about others. | Individuals who score high on this construct tend to believe in other people's benevolent intentions.<br><br>Individuals who score low on this construct show difficulty in holding positive beliefs about the intentions of others. Very low trust values can be associated with cynicism. | (Costa et al. 1991, Johnson 2014, Soto and John 2017) |
| Variety-seeking | The will of an individual to continuously look for different, new, and innovative experiences, generating a sense of excitement and enthusiasm. | Individuals who score high on this construct are typically inclined to feel a strong emotional state of excitement when discovering and living new experiences. Very high scorers may have tendencies toward dangerous behaviors and addictions.<br><br>Individuals who score low on this construct usually do not experience new situations with enthusiasm. Very low scorers may show a tendency to routinary habits. | (Bommelje et al. 2003, Le Bon et al. 2004, Kan et al. 2022) |

The following table presents some examples of keywords associated with each psychological construct, which we used for the text mining analysis.

**Table B.2:** examples of selected keywords for the psychological constructs

| Dimension label | Keywords |
|---|---|
| Achievement-Striving | "achievement", "accomplishment", "striving", "excellence", "target", "results", "result" |
| Adventurousness | "adventure", "action", "actions" |
| Ambition/Drive | "ambition", "success", "successful", "exceeding" |
| Competence | "proven competence", "knowledge", "profession", "qualified", |
| Complexity | "complex", "difficulty", "complicated", "intriguing", "critical" |
| Comprehension | "comprehend", "understanding", "mastery of language", "perception" |
| Construction/ Mechanical Interests | "construction", "mechanical interest", "construction interest", "mechanical", "mechanical interests", "construction interests" |
| Cooperation | "cooperation", "collaborative", "collaborate", "collaboration", "collaborating" |
| Creativity/Originality | "creativity", "creating", "create", "originality", "inventing" |
| Dutifulness | "duty", "duties", "dutifulness", "dutifully", "dutiable" |
| Efficiency | "efficient", “efficiency” |
| Initiative | "take initiative", "initiatives", "proactive" |
| Interest in Environmentalism | "environment", "interest in environmentalism", “sustainability” |
| Language Mastery | "language mastery", "master of language", "rich vocabulary", "vocabulary", "large vocabulary", "language-mastery", "speak fluently", "fluent language", "fluency in language", "public speaking", "communication", "communicate", "communicating" |
| Leadership | "leader", "leads", "lead", "leading", "coach", "accountability", "coordinate" |
| Love of Learning | "love for learning", "love of learning", "learning", "grow*" |
| Methodicalness | "methodology", "accurate", "accuracy", "precise", "precision" |
| Orderliness | "order", “tidy” |
| Planfulness | "plan", "plans", "planning", "pianification", "timeline", "deadline", "schedule", "scheduling", "schedules" |
| Problem-solving | "problem-solving", "problem solving", "out of the box", "out-of-the-box", "lateral thinking", "lateral-thinking", "solutions", "problem solver", "troubleshoot*" |
| Prudence | "prudence", "safe" |
| Purposefulness | "purpose", "tasks", "task", "objective", "objectives", "goal", "goals" |
| Quickness | "quick", "quickly", "prompt", "promptly", "fast", "fastly", "speedy" |
| Responsibility | "responsible", "responsibilities", “accountable”, “accountability” |
| Security | "security", "rely", "reliable", "consistent" |
| Teamwork/ Citizenship | "teamwork", "loyalty", "citizenship", "partnering", "cross-functional", "cross-function", “team” |
| Trust | "trust", "committed", "commitment", "commit", "commits" |
| Variety-seeking | "variety seeking", "variety-seeking", "changing", "variety", "innovation" |